\documentclass[11pt,a4paper]{article}
\usepackage[hyperref]{acl2019}
\usepackage{times}
\usepackage{booktabs}
\usepackage{latexsym}
\usepackage{todonotes}
\usepackage{graphicx}
\usepackage{amsmath}
\usepackage{url}

\aclfinalcopy 
\def\aclpaperid{1278} 

\usepackage{tabularx}
\newcolumntype{L}[1]{>{\raggedright\arraybackslash}p{#1}}

\title{Putting words in context: \\ LSTM language models and lexical ambiguity}

\author{Laura Aina    \hspace{4ex} Kristina Gulordava\hspace{5ex}
	Gemma Boleda \\ Universitat Pompeu Fabra \\ Barcelona, Spain \\ {\tt \{firstname.lastname\}@upf.edu}	}
\date{}

\begin{document}
\maketitle
\begin{abstract}

In neural network models of language, words are commonly represented using context-invariant representations (word embeddings) which are then put in context in the hidden layers.
Since words are often ambiguous, representing the contextually relevant information is not trivial.
We investigate how an LSTM language model deals with lexical ambiguity in English, designing a method to probe its hidden representations for lexical and contextual information about words.
We find that both types of information are represented to a large extent, but also that there is room for improvement for contextual information.
\end{abstract}

\section{Introduction}

\begin{table*}[t]
	\footnotesize{
	\begin{tabular}{L{3.8cm}L{2.5cm}L{2.5cm}L{2.5cm}L{2.5cm}}
			Examples &  LexSub & $\mathbf{w}$ \textsc{nn} & $\mathbf{s}$ \textsc{nn} & $\mathbf{w\&s}$ \textsc{nn} \\ \toprule
		(1) \textit{\dots I clapped her shoulder to \textbf{show} I was not laughing at her\dots }
		& demonstrate, display, indicate, prove, clarify 
		& demonstrate, exhibit, indicate, offer, reveal 
		& indicate, demonstrate, suggest, prove, 
		& indicate, demonstrate, prove, ensure, suggest 
		\\ \midrule
		(2) 
\textit{\dots The \textbf{show} [\dots] revolutionized the way America cooks and eats\dots} 
		& program, series, broadcast, presentation 
		& demonstrate, exhibit, indicate, offer, reveal 
		& series, program, production, miniseries, trilogy 
		& series, program, production, broadcast \\
		 \midrule
		(3) \textit{\dots The inauguration of Dubai Internet City coincides with the opening of an annual IT \textbf{show} in Dubai\dots}. 
		& exhibition, conference, convention, demonstration 
		& demonstrate, exhibit, indicate, offer, reveal 
		& conference, event, convention, symposium, exhibition 
		& conference, event, exhibition, symposium, convention 
		\\ \bottomrule
	\end{tabular}}
	\caption{Examples from the LexSub dataset \cite{kremer2014} and nearest neighbors for target representations.}
	\label{tab:lexsubexamples}
	\vspace{-0.2cm}
\end{table*}


In language, a word can contribute a very different meaning to an utterance depending on the context, a phenomenon known as lexical ambiguity \cite{cruse1986, small2013}.
This variation is pervasive and involves both morphosyntactic and semantic aspects.
For instance, in the examples in Table~\ref{tab:lexsubexamples}, \textit{show} is used as a verb in Ex.~(1), and as a noun in Ex.~(2-3), in a paradigmatic case of morphosyntactic ambiguity in English.
Instead, the difference between Ex.~(2) and~(3) is semantic in nature, with \textit{show} denoting a TV program and an exhibition, respectively.
Semantic ambiguity covers a broad spectrum of phenomena, ranging from quite distinct word senses (e.g.\ \textit{mouse} as animal or computer device) to more subtle lexical modulation (e.g.~\textit{visit a city / an aunt / a doctor};~\citealp{cruse1986}).
This paper investigates how deep learning models of language, and in particular Long Short-Term Memory Networks (LSTMs) trained on Language Modeling, deal with lexical ambiguity.%
\footnote{Code at: \url{https://github.com/amore-upf/LSTM_ambiguity}}

In neural network models of language, words in a sentence are commonly represented through word-level representations that do not change across contexts, that is, ``static'' word embeddings.
These are then passed to further processing layers, such as the hidden layers in a recurrent neural network (RNN).
Akin to classic distributional semantics~\cite{erk2012}, word embeddings are formed as an abstraction over the various uses of words in the training data.
For this reason, they are apt to represent context-invariant information about a word ---its \textbf{lexical} information--- but not the contribution of a word in a particular context ---its \textbf{contextual} information~\cite{erk2010}.
Indeed, word embeddings subsume information relative to various senses of a word~(e.g., \textit{mouse} is close to words from both the animal and computer domain;~\citealp{camacho2018}).

Classic distributional semantics attempted to do composition to account for contextual effects, but it was in general unable to go beyond short phrases~\cite{baroni2013composition}; newer-generation neural network models have supposed a big step forward, as they can natively do composition~\cite{westera-boleda-2019-dont}.
In particular, the hidden layer activations in an RNN can be seen as \textit{putting words in context}, as they combine the word embedding with information coming from the context (the adjacent hidden states).
The empirical success of RNN models, and in particular LSTM architectures, at fundamental tasks like Language Modeling~\cite{jozefowicz2015} suggests that they are indeed capturing relevant contextual properties.
Moreover, contextualized representations derived from such models
have been shown to be very informative as input for lexical disambiguation tasks~\cite[e.g.][]{melamud2016,peters2018}.

We here present a method to probe the extent to which the hidden layers of an LSTM language trained on English data represent lexical and contextual information about words, in order to investigate how the model copes with lexical ambiguity.
Our work follows a recent strand of research that purport to identify what linguistic properties deep learning models are able to capture~\cite[a.o.]{linzen2016,adi2016,gulordava2018,conneau2018,hupkes2018}.
We train diagnostic models on the tasks of retrieving the embedding of a word and a representation of its contextual meaning, respectively ---the latter obtained from a Lexical Substitution dataset~\cite{kremer2014}.
Our results suggest that LSTM language models heavily rely on the lexical information in the word embeddings, at the expense of contextually relevant information.
Although further analysis is necessary, this suggests that there is still much room for improvement to account for contextual meanings.
Finally, we show that the hidden states used to predict a word -- as opposed to those that receive it as input -- display a bias towards contextual information.


\section{Method}
\label{sec:method}

\paragraph{Language model.}
As our base model, we employ a word-level bidirectional LSTM~\cite{schuster1997,hochreiter1997} language model (henceforth, LM) with three hidden layers.
Each input word at timestep $t$ is represented through its word embedding $\mathbf{w}_t$; this is fed to both a forward and a backward stacked LSTMs, which process the sequence left-to-right and right-to-left, respectively~(Eqs.~(\ref{eq:layer1}-\ref{eq:layeri}) describe the forward LSTM).
To predict the word at $t$, we obtain output weights by summing the activations of the last hidden layers of the forward and backward LSTMs at timesteps $t-1$ and $t+1$, respectively, and applying a linear transformation followed by softmax (Eq.\ \ref{eq:output}, where $L$ is the number of hidden layers). 
Thus, a word is predicted using both its left and right context jointly, akin to the \textit{context2vec} architecture~\cite{melamud2016} but differently from, e.g.,~the BiLSTM architecture used for ELMo~\cite{peters2018}.
\begin{align}
\label{eq:layer1}
\mathbf{h}_{t}^1 = \text{LSTM}^1(\mathbf{w}_t, \mathbf{h}_{t-1}^1) \\
\label{eq:layeri}
\mathbf{h}_{t}^i = \text{LSTM}^i(\mathbf{h}_{t}^{i - 1}, \mathbf{h}_{t-1}^i)\\
\label{eq:output}
 \mathbf{o}_t = softmax(f (\overrightarrow{\mathbf{h}}^L_{t -1} + \overleftarrow{\mathbf{h}}^L_{t +1}))
\end{align}

\noindent We train the LM on the concatenation of English text data from a Wikipedia dump\footnote{From 2018/01/03, \url{https://dumps.wikimedia.org/enwiki/}}, the British National Corpus \cite{leech1992}, and the UkWaC corpus \cite{ferraresi2008}.\footnote{50M tokens from each corpus, in total 150M (train/valid/test: 80/10/10\%); vocabulary size: 50K.}
More details about the training setup are specified in Appendix~\ref{app:lm}.
The model achieves satisfying performances on test data~(perplexity: 18.06).

\begin{figure}
	\centering
	\includegraphics[width=6cm]{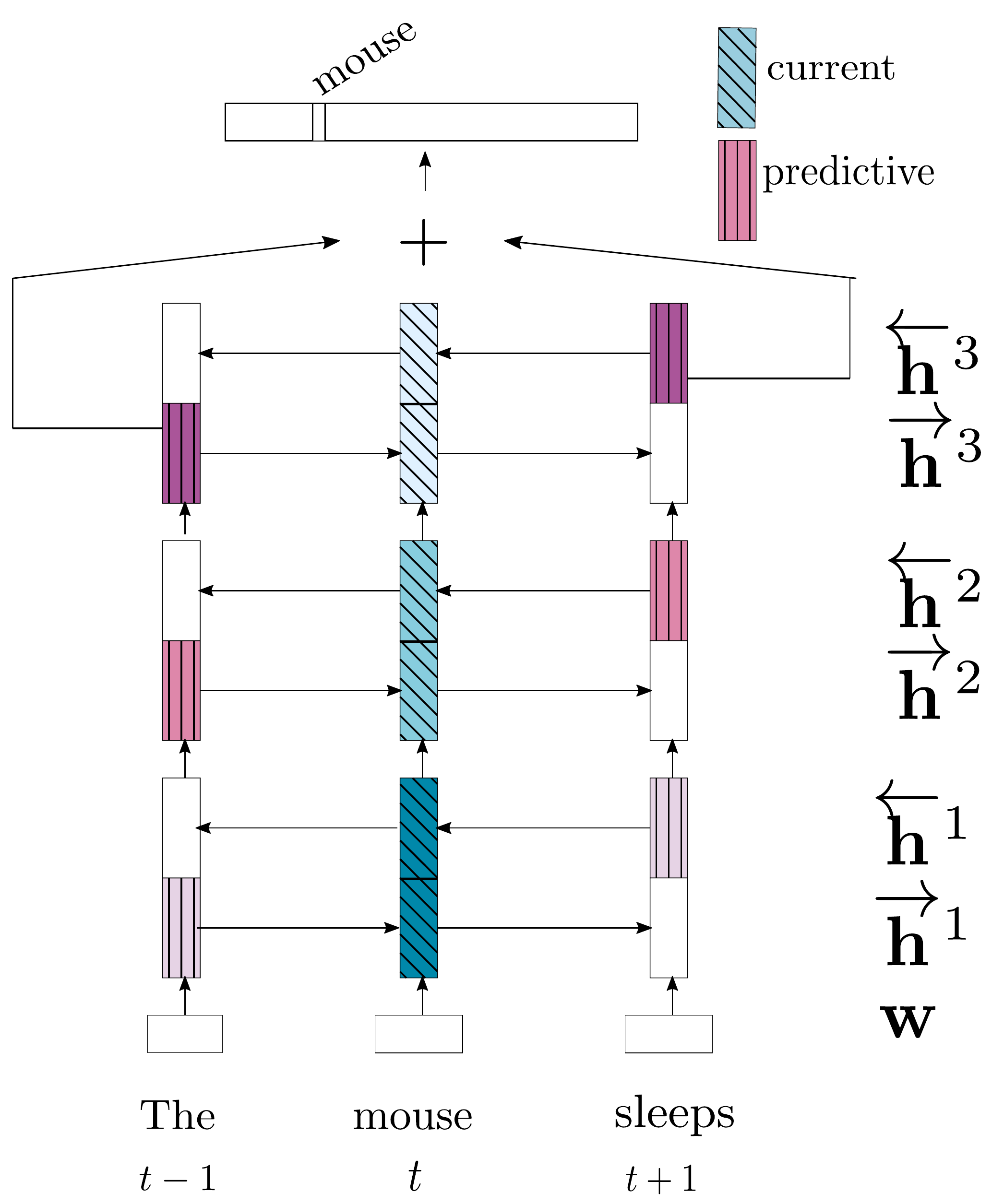}
	\caption{Language model and extracted representations. The different shades across layers reflect the different performances in the probe tasks (darker = higher)}
	\label{fig:model}
	\vspace{-0.4cm}
\end{figure}

For our analyses, we deploy the trained LM on a text sequence and extract the following activations of each hidden layer; Eq.~(\ref{eq:inputprobe1}) and Fig.~\ref{fig:model}.
\begin{align}
\label{eq:inputprobe1}
\{ \overrightarrow{\mathbf{h}}^i_{t}| i \leq L\} \cup \{ \overleftarrow{\mathbf{h}}^i_{t}| i \leq L\}\\
\label{eq:inputprobe2}
\mathbf{h}_t^i = [ \overrightarrow{\mathbf{h}}^i_{t} ; \overleftarrow{\mathbf{h}}^i_{t}]\\
\label{eq:predictive}
\mathbf{h}_{t\pm1}^i = [ \overrightarrow{\mathbf{h}}^i_{t -1} ; \overleftarrow{\mathbf{h}}^i_{t+1} ]
\end{align}
At timestep $t$, for each layer, we concatenate the forward and backward hidden states; Eq.~(\ref{eq:inputprobe2}).
We refer to these vectors as \textbf{current hidden states}.
As they are obtained processing the word at $t$ as input and combining it with information from the context, we can expect them to capture the relevant contribution of such word (e.g., in Fig.~\ref{fig:model} the mouse-as-animal sense).
As a comparison, we also extract activations obtained 
by processing the text sequence up to $t-1$ and $t+1$ in the forward and backward LSTM, respectively, hence excluding the word at $t$. 
We concatenate the forward and backward states of each layer;  Eq.~(\ref{eq:predictive}).
While these activations do not receive the word at $t$ as input, they are relevant because they are used to predict that word as output.
We refer to them as \textbf{predictive hidden states}. 
These may capture some aspects of the word (e.g., in Fig.~\ref{fig:model}, that it is a noun and denotes something animate), but are likely to be less accurate than the current states, since they do not observe the actual word.

\paragraph{Probe tasks.}

We aim to assess to what extent the hidden states in the LM carry over the lexical and context-invariant information in the input word embedding, and how much they instead represent the contextual meaning of the word.
To this end, we rely on vector representations of lexical and contextual word information.
As for the former, we can directly use the word embeddings of the LM ($\textbf{w}$); it is instead more challenging to find a representation of the contextual meaning. 

Our solution is to use Lexical Substitution data \cite{mccarthy2009}
and, in particular, the large dataset by~\citealp{kremer2014}~(henceforth, LexSub; see Table~\ref{tab:lexsubexamples}). In this dataset, words in context (up to 3 sentences) are annotated with a set of paraphrases given by human subjects.
Since contextual substitutes reflect differences among uses of a word (for instance, \textit{demonstrate} paraphrases \textit{show} in a context like Ex.~(1), but not in Ex.~(2)), this type of data is often used as an evaluation benchmark for contextual representations of words~(e.g.,~\citealp{erk2008,melamud2016,GarSoler2019}).
We leverage LexSub to build proxies for ground-truth representations of the contextual meaning of words.
We define two types of representations, inspired by previous work that proposed simple vector operations to combine word representations \cite[a.o.]{mitchell2010,thater2011}:
the average embedding of the substitute words (henceforth, $\mathbf{s}$), and the average embedding of the union of the substitute words and the target word ~($\mathbf{w\&s}$).
As Table~\ref{tab:lexsubexamples} qualitatively shows, the resulting representations tend to be close to the substitute words and reflect the contextual nuance conveyed by the word; in the case of $\mathbf{w\&s}$, they also retain a strong similarity to the embedding of the target word.\footnote{These vectors are close to the related word embedding (0.45 and 0.66 mean cosine, see Table~\ref{table:results}, row $\mathbf{w}_t$), but also different from it: on average, $\mathbf{s}$ and $\mathbf{w\&s}$ share 17 and 25\% of the top-10 neighbors with $\mathbf{w}$, respectively (statistics from training data, excluding the word itself from neighbors).}

We frame our analyses as supervised probe tasks: a diagnostic model learns to ``retrieve'' word representations out of the hidden states; the rate of success of the model is taken to measure the amount of information relevant to the task that its input contains.
Given current or predictive states as inputs, we define three diagnostic tasks:
\begin{itemize}\vspace{-0.2cm}
	\item[-] \textsc{Word}: predict $\mathbf{w}$ \vspace{-0.2cm}
	\item[-] \textsc{Sub}: predict $\mathbf{s}$ \vspace{-0.2cm}
	\item[-]\textsc{Word\&Sub}: predict $\mathbf{w\&s}$
\end{itemize} 
\noindent The \textsc{Word} task is related to the probe tasks introduced in~\citet{adi2016} and~\citet{conneau2018}, which, given a hidden state, require to predict the words that a sentence encoder has processed as input. Note that, while these authors predict words by their discrete index, we are predicting the complete multi-dimensional embedding of the word. Our test quantifies not only whether the model is tracking the identity of the input word, but also how much of its information it retains.

We train distinct probe models for each task and type of input ($\mathbf{i}$; e.g., current hidden state at layer 1). 
A model consists of a non-linear transformation from an input vector $\mathbf{i}$ (extracted from the LM) to a vector with the dimensionality of the word embeddings (Eq.~\ref{eq:predict}, where $\mathbf{\hat{r}}$ is one of $\mathbf{\hat{w}}$, $\mathbf{\hat{s}}$, $\mathbf{\hat{w\&s}}$ for \textsc{Word}, \textsc{Sub}, and \textsc{Word\&Sub} tasks, respectively).
The models are trained through max-margin loss, optimizing the cosine similarity between $\mathbf{\hat{r}}$ and the target representation against the similarities between $\mathbf{\hat{r}}$ and 5 negative samples (details in Appendix~\ref{app:dm}).
\begin{equation}
\label{eq:predict}
	\mathbf{\hat{r}} = tanh (W \ \mathbf{i} + b )\vspace{-0.1cm}
\end{equation}

We adapt the LexSub data to our setup as follows.
Since substitutes are provided in their lemmatized form, we only consider datapoints where the word form is identical to the lemma so as to exclude effects due to morphosyntax (e.g., asking the models to recover \textit{play} when they observe \textit{played}).\footnote{We also exclude substitutes that are multi-word expressions and the datapoints involving words that are part of a compound (e.g., \textit{fast} in \textit{fast-growing}).}
We require that at least 5 substitutes per datapoint are in the LM vocabulary to ensure quality in the target representations.
LexSub data come with a validation/test split; since we need training data, we create a new random partitioning into train/valid/test (70/10/20\%, with no overlapping contexts among splits).
The final data consist of 4.7K/0.7K/1.3K datapoints for train/valid/test.


\section{Results}
\label{sec:results}

The results of the probe tasks on test data are presented in Table \ref{table:results}. 
We report the mean and standard deviation of the \textbf{cosine} similarity between the output representations ($\mathbf{\hat{w}}$, $\mathbf{\hat{s}}$, $\mathbf{\hat{w\&s}}$) and the target ones ($\mathbf{w}$, $\mathbf{s}$, $\mathbf{w\&s}$).
This evaluates the degree to which the word representations can be retrieved from the hidden states.
For comparison, we also report the cosine scores between the targets and two baseline representations: the word embedding itself and the average of word embeddings of a 10-word window around the target word~($avg_{ctxt}$).\footnote{We exclude out-of-vocabulary words and punctuation.}
Overall, the models do better than these unsupervised baselines, with exceptions.\footnote{The first cell is 1 as it involves the same representation.}

\begin{table}
	\centering
	\begin{tabular}{lccc}\toprule
		
		{\small input} & {\small \textbf{\textsc{Word}}} & {\small \textbf{\textsc{Sub}}} & {\small \textbf{\textsc{Word\&Sub}}} \\ \midrule
		$\mathbf{w}_t $ & 1  & .45 ($\pm .14$) &  .66 ($\pm .09$) \\
		{\small $avg_{ctxt}$}	&  .35 ($\pm .10$) & .16 ($\pm .11$) & .24 ($\pm .12$) \\\midrule
					
		$\mathbf{h}_{t}^1$& \textbf{.84} ($\pm .2$)& \textbf{.61} ($\pm .14$) & \textbf{.71} ($\pm .11$)\\
		$\mathbf{h}_{t}^2$& .74 ($\pm .12$) &   .60 ($\pm .13$) &.69 ($\pm .11$)\\ 
		$\mathbf{h}_{t}^3$& .64 ($\pm .12$) & .58 ($\pm .13$)& .65 ($\pm .11$)\\
		\midrule
		
		$\mathbf{h}_{t\pm1}^1$ &  .25 ($\pm .16$) & .36 ($\pm .16$) &  .38 ($\pm .16$)\\
		$\mathbf{h}_{t\pm1}^2$ & .27 ($\pm .16$)& .39 ($\pm .16$) & .41 ($\pm .16$)\\
		$\mathbf{h}_{t\pm1}^3$ & \textbf{.29} ($\pm .15$)& \textbf{.41} ($\pm .16$) & \textbf{.43} ($\pm .16$) \\
		\bottomrule 	
	\end{tabular}
	\caption{Results of probe tasks for current ($\mathbf{h}^i_t$) and predictive ($\mathbf{h}_{t\pm1}^i$) hidden states.}
	\label{table:results}
	\vspace{-0.3cm}
\end{table}

\begin{figure}[]
	\centering
	\vspace{-0.4cm}
	\includegraphics[width=5cm]{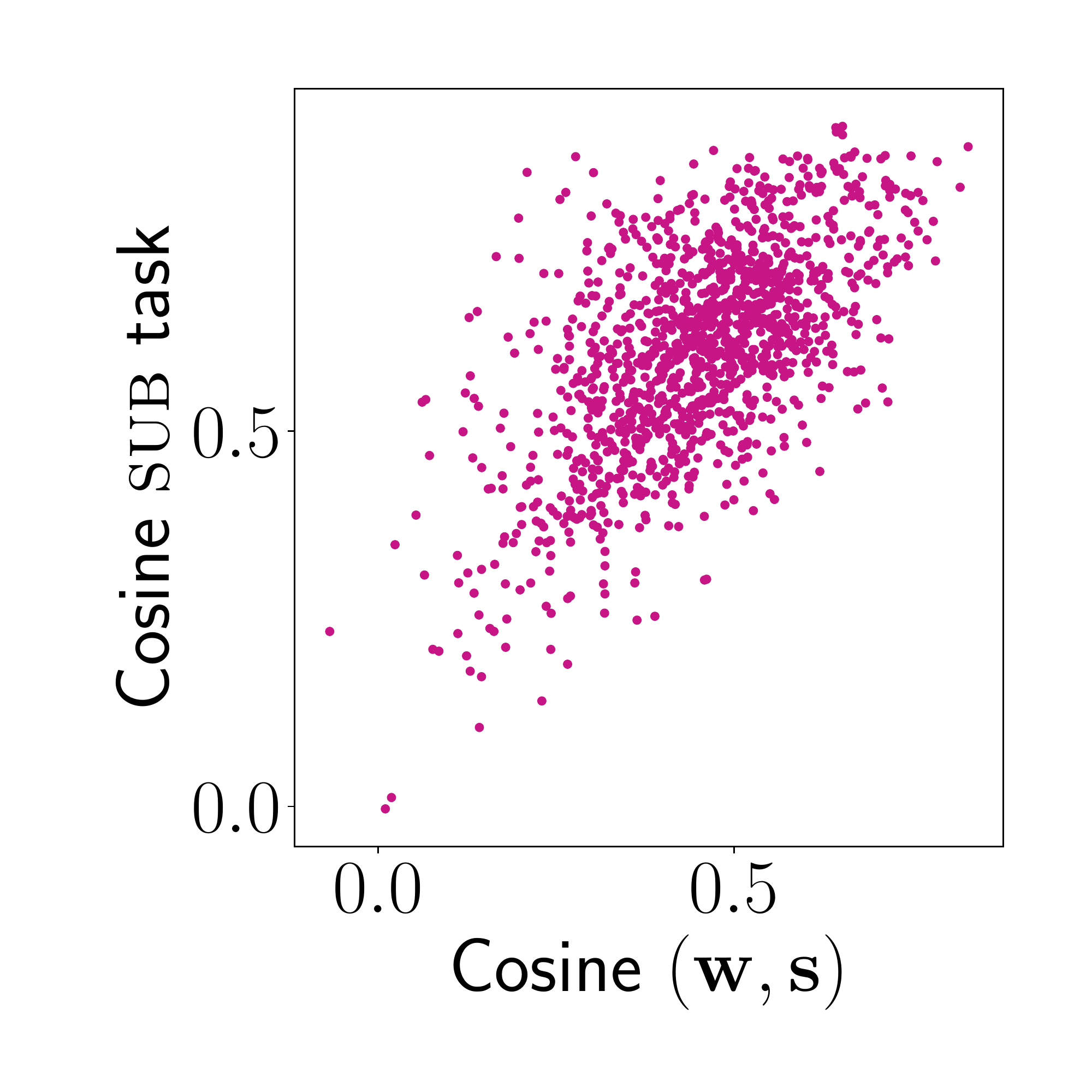}
	\vspace{-0.4cm}
	\caption{Similarity of lexical and contextual vector ($\mathbf{w}$ - $\mathbf{s}$) vs.\ similarity of target and prediction in \textsc{Sub} for $\mathbf{h}_{t}^1$.}
	\label{fig:corr}
	\vspace{-0.5cm}
\end{figure}

\begin{table*}\center
	\small{
		\begin{tabular}{L{3.7cm}L{2.9cm}L{2cm}L{2.3cm}L{3.1cm}}
			Context &  LexSub & \textsc{Word}: $\mathbf{\hat{w}}$ \textsc{nn} & \textsc{Sub}: $\mathbf{\hat{s}}$ \textsc{nn} & \textsc{Word\&Sub}: $\mathbf{w\hat{\&}s}$ \textsc{nn} \\ \toprule
			(4) ... \textit{The financial-services company will pay 0.82 share for each Williams \textbf{share}} ... & stock, dividend, interest, stake, unit &  stake, owe, discuss, coincide, reside & 
			portion, amount, percentage, fraction 
			& stake, percentage, portion, spend, proportion  \\ \midrule
			(5) 
 ...  \textit{Sony's effort to hire producers Jon Peters and Peter Guber to run the \textbf{studio}}...  
& business, company, facility, film, lot & lab, troupe, classroom, apartment, booth & room, gallery, troupe, journal, house & room, troupe, lab, audience, department \\ \midrule
			(6) 
			\textit{...  I had [...] told her that we needed other \textbf{company} than our own  ...
			} 
			& friend, acquaintance, visitor, accompaniment, associate & retailer, trader, firm, maker, supplier & firm, corporation, organisation, conglomerate, retailer & corporation, firm, conglomerate, retailer, organisation \\ 
			\bottomrule
	\end{tabular}}
	\caption{Examples with nearest neighbours of the representations predicted in the first current hidden layer.}
	\label{table:examples}
	\vspace{-0.3cm}
\end{table*}

\paragraph{Current hidden states.}
Both lexical and contextual representations can be retrieved from the current hidden states ($\mathbf{h}^i_t$) to a large extent (cosines .58-.84), but
retrieving the former is much easier than the latter (.64-.84 vs.\ .58-71).
This suggests that the information in the word embedding is better represented in the hidden states than the contextually relevant one.
In all three tasks, performance degrades closer to the output layer (from $\mathbf{h}_{t}^1$ to $\mathbf{h}_{t}^3$), but the effect is more pronounced for the \textsc{Word} task (84/.74/.64).
Word embeddings are part of the input to the hidden state, and the transformation learned for this task can be seen as a decoder in an auto-encoder, reconstructing the original input; the further the hidden layer is from the input, the more complex the function is to reverse-engineer. 
Crucially, the high performance at reconstructing the word embedding suggests that lexical information is retained in the hidden layers, possibly including also contextually irrelevant information (e.g., in Ex.~(4) in Table~\ref{table:examples} $\mathbf{\hat{w}}$ is close to verbs, even if \textit{share} is here a noun).

Contextual information ($\mathbf{s}$ and $\mathbf{w\&s}$) seems to be more stable across processing layers, although overall less present (cf.\ lower results).
Table \ref{table:examples} reports one example where the learned model displays relevant contextual aspects (Ex.\ (4), \textit{share}) and one where it does not (Ex.\ (5), \textit{studio}).
Qualitative analysis shows that morphosyntactic ambiguity (e.g., \textit{share} as a noun vs.~verb) is more easily discriminated, while semantic distinctions pose more challenges (e.g., \textit{studio} as a room vs.~company).
This is not surprising, since the former tends to correlate with clearer contextual cues.
Furthermore, we find that the more the contextual representation is aligned to the lexical one, the easier it is to retrieve the former from the hidden states~(e.g.,  correlation $\cos(\mathbf{w}, \mathbf{s})$ - $\cos(\mathbf{\hat{s}}, \mathbf{s})$, for $\mathbf{h}_{t}^1$: Pearson's~$\rho= .62^{***}$; Fig.~\ref{fig:corr}): that is, it is harder to resolve lexical ambiguity when the contextual meaning is less represented in the word embedding~(e.g., less frequent uses).
This suggests that the LM heavily relies on the information in the word embedding, making it challenging to diverge from it when contextually relevant (see Ex.~(6) in Table~\ref{table:examples}).

\paragraph{Current vs.~predictive hidden states.} 
The predictive hidden states are obtained without observing the target word; hence, recovering word information is considerably harder than for current states.
Indeed, we observe worse results in this condition~(e.g.,~below $avg_{ctxt}$ in the \textsc{word} task); we also observe two patterns that are opposite to those observed for current states, which shed light on how LSTM LMs track word information. 

For predictive states, results improve closer to the output (from layer 1 to 3; they instead degrade for current states).
We link this to the double objective that a LM has when it comes to word information: to integrate a word passed as input, and to predict one as output.
Our results suggest that the hidden states keep track of information for both words, but lower layers focus more on the processing of the input and higher ones on the predictive aspect~(see Fig.~\ref{fig:model}).
This is in line with previous work showing that activations close to the output tend to be task-specific~\cite{liu2019}.

Moreover, from predictive states, it is easier to retrieve contextual than lexical representations (.41/.43 vs.\ .29; the opposite was true for current states).
Our hypothesis is that this is due to a combination of two factors.
On the one hand, predictive states are based solely on contextual information, which highlights only certain aspects of a word; for instance, the context of Ex.~(2) in Table~\ref{tab:lexsubexamples} clearly signals that a noun is expected, and the predictive states in a LM should be sensitive to this kind of cue, as it affects the probability distribution over words.
On the other hand, lexical representations are underspecified; for instance, the word embedding for \textit{show} abstracts over both verbal and nominal uses of the word.
Thus, it makes sense that the predictive state does not capture contextually irrelevant aspects of the word embedding, unlike the current state (note however that, as stated above, the overall performance of the current state is better, because it has access to the word actually produced).


\section{Future work}
\label{sec:conc}

We introduced a method to study how deep learning models of language deal with lexical ambiguity.
Though we focused on LSTM LMs for English, this method can be applied to other architectures, objective tasks, and languages; possibilities to explore in future work. 
We also plan to carry out further analyses aimed at individuating factors that challenge the resolution of lexical ambiguity~(e.g., morphosyntactic vs.~semantic ambiguity, frequency of a word or sense, figurative uses), as well as clarifying the interaction between prediction and processing of words within neural LMs.


\section*{Acknowledgements}
This project has received funding from the European Research Council
(ERC) under the European Union's Horizon 2020 research and innovation
programme (grant agreement No 715154),
and from the Ram\'on y Cajal programme (grant RYC-2015-18907).
We gratefully acknowledge the support of NVIDIA Corporation with the donation of GPUs used for this research, and the computer resources at CTE-POWER and the technical support provided by Barcelona Supercomputing Center (RES-FI-2018-3-0034).
This paper reflects the authors' view only, and the EU is not
responsible for any use that may be made of the information it
contains.\begin{flushright}
	\includegraphics[width=0.6cm]{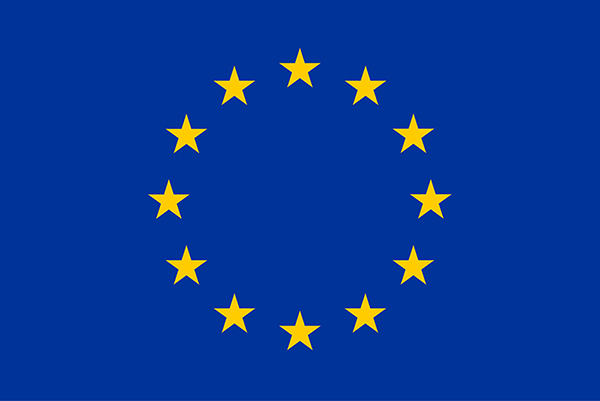}  
	\includegraphics[width=0.6cm]{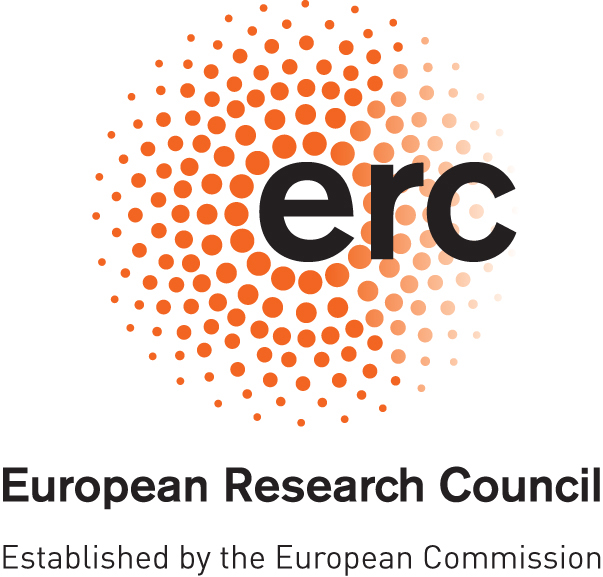} 
\end{flushright}

\bibliography{references}
\bibliographystyle{acl_natbib}



\appendix
\section{Appendix}
	
\subsection{Language model}
\label{app:lm}
The hidden layers are of sizes 600/600/300 respectively, while the word embeddings are of size 300.
The language model was trained optimizing the log-likelihood of a target word given its surrounding context, with stochastic gradient descent for 20 epochs with decaying learning rate using Adam optimiser \cite{kingma2014adam}. The initial learning rate was 0.0005 for batch size of 32. Dropout was set to 0.2 and applied to the input embedding,
and the outputs of the LSTM layers.
At training time, the text data is fed to the model in sequences of 100 tokens. 

\subsection{Diagnostic models}
\label{app:dm}
We train separate models for each combination of task and input type.
Each model consist of a linear transformation and a $tahn$ non-linearity, trained using Cosine Embedding Loss (PyTorch 0.4,  \citealp{paszke2017automatic}) and Adam optimiser, with early stopping based on validation loss. 
We carried out hyperparameter search based on validation loss for each of the model types in order to set batch size and initial learning rate. 
We report the final settings for each combination of input and task in Table \ref{table:hp}.

At training time, for each positive target word, we obtain 5 negative targets by sampling words from the frequency quartile of the postive target (frequency is computed on the training corpus of the language model).
We always exclude the target word, as well as the substitute words in the \textsc{sub} and \textsc{word\&sub} conditions, from the negative samples.
Given the input vector, we maximize the margin of the resulting output vector $\hat{r}$ to the embeddings of the negative samples ($i = -1$), and minimize the distance of the output vector to the target representation of the positive instance ($i = 1$; Eq.~\ref{eq:loss}). 

\begin{equation}
L(\hat{r}, r, i) = \\
\begin{cases}
1 - cos(\hat{r}, r) \hspace{2cm} \text{if } i = 1 \\
\\
\hspace{4cm} \text{if } i = - 1  \\
max(0, cos(\hat{r}, r) - \text{margin}) 
\end{cases}
\label{eq:loss} 
\end{equation}

\noindent At each training epoch, new negative instances are sampled, and the data is shuffled.

	\begin{table}\center
		\footnotesize{
		\begin{tabular}{L{0.4cm}ccc}\toprule
			input & \textsc{Word} & \textsc{sub} & \textsc{Word\&sub} \\ \midrule
			
			$\mathbf{h}_{t}^1$& $16, 5\times10^{-5}$ & $32, 1\times10^{-4}$  & $32, 5\times10^{-5}$\\
			$\mathbf{h}_{t}^2$ & $ 16, 5\times10^{-5}$ & $  64,5\times10^{-4} $& $ 64, 5\times10^{-4}$\\ 
			$\mathbf{h}_{t}^3$ & $ 16, 5\times10^{-5} $ & $ 128,5\times10^{-4}$ & $ 16, 5\times10^{-5}$ \\
			\midrule
			$\mathbf{h}_{t\pm1}^1$ & $  128, 1\times10^{-3} $ & $ 128, 1\times10^{-3} $& $  128, 5\times10^{-4}$\\
			$\mathbf{h}_{t\pm1}^2$ & $ 16, 1\times10^{-4} $&  $  64, 5\times10^{-4}$&  $ 16, 5\times10^{-4}$\\
			$\mathbf{h}_{t\pm1}^3$ & $ 128, 1\times10^{-3}$ & $ 16, 1\times10^{-4} $  & $ 128, 5\times10^{-4}$ \\
\bottomrule 	
		\end{tabular}
		\caption{Hyperparameter settings in the diagnostic models (batch size, initial learning rate)}
		\label{table:hp}}
	\end{table}


\end{document}